\documentclass[graybox]{svmult}

\usepackage[l2tabu,orthodox]{nag}

\usepackage[square,numbers]{natbib}
\usepackage[ pdftex,colorlinks]{hyperref} %

\usepackage{layouts} %

\usepackage[printonlyused]{acronym}
\usepackage[binary-units]{siunitx}
\usepackage[all]{nowidow}

\usepackage[dvipsnames, table]{xcolor}

\usepackage{lipsum}

\usepackage{verbatim}

\usepackage{mathtools}

\usepackage{subcaption}

\usepackage{sidecap}

\usepackage{pdflscape}

\usepackage{pifont}
\usepackage{soul}
\usepackage{xcolor}
\DeclareRobustCommand{\ctext}[2]{\sethlcolor{#1}\hl{#2}}
\soulregister{\ctext}{2}

\usepackage[pdftex]{graphicx}
\usepackage{epstopdf}

\usepackage{import}

\usepackage{float}

\usepackage[font=footnotesize]{caption}

\graphicspath{{./latexGoodPractices/}}

\usepackage[percent]{overpic}

\usepackage{copyrightbox}

\usepackage{booktabs}

\usepackage{tabu}

\usepackage{supertabular}

\usepackage{rotating,tabularx}

\usepackage{makecell}

\usepackage{amssymb,amsfonts,amsmath,amscd}

\usepackage{bm}

\newcommand{\bbm}{\begin{bmatrix}}
\newcommand{\ebm}{\end{bmatrix}}

\DeclareFontFamily{U}{mathx}{\hyphenchar\font45}
\DeclareFontShape{U}{mathx}{m}{n}{
      <5> <6> <7> <8> <9> <10>
      <10.95> <12> <14.4> <17.28> <20.74> <24.88>
      mathx10
      }{}
\DeclareSymbolFont{mathx}{U}{mathx}{m}{n}
\DeclareFontSubstitution{U}{mathx}{m}{n}
\DeclareMathAccent{\widecheck}{0}{mathx}{"71}
\DeclareMathAccent{\wideparen}{0}{mathx}{"75}

\DeclareMathAlphabet{\mathcal}{OMS}{cmsy}{m}{n}

\acrodef{ICP}{Iterative Closest Point}
\acrodef{L2}{squared distance}
\acrodef{NDT}{Normal Distribution Transformation}
\acrodef{GICP}{Generalized-ICP} 
\acrodef{SLAM}{Simultaneous Localization and Mapping}
\acrodef{GPS}{Global Positioning System}
\acrodef{ROS}{Robotic Operating System}
\acrodef{IMU}{Inertial Measurement Unit}
\acrodef{EKF}{Extended Kallman Filter}
\acrodef{GNSS}{Global Navigation Satellite System}
\acrodef{GTLS-ICP}{Generalized Total-Least-Squares ICP}
\acrodef{BFGS}{Broyden-Fletcher-Goldfarb-Shanno}
\acrodef{DoF}{Degrees of Freedom}
\acrodef{ALS}{Airborne Lidar Scanning}
\acrodef{RTK}{Real-Time Kinematic}
\acrodef{ENU}{East-North-Up}
\acrodef{NDT}{Normal Distributions Transform}

\usepackage{mathptmx}       %
\usepackage{helvet}         %
\usepackage{courier}        %
\usepackage{type1cm}        %
\usepackage{makeidx}         %
\usepackage{graphicx}        %
\usepackage{multicol}        %
\usepackage[bottom]{footmisc}%

\usepackage[normalem]{ulem}  %

\makeindex             %

\begin{document}
\newlength\myboxwidth
\setlength{\myboxwidth}{\dimexpr\textwidth-2\fboxsep}

\authorrunning{P. Babin, P. Dandurand, V. Kubelka, P. Gigu\`{e}re and F. Pomerleau}

\title*{Large-scale 3D Mapping of Subarctic Forests}
\author{Philippe Babin, Philippe Dandurand, Vladim\'{i}r Kubelka, Philippe Gigu\`{e}re and Fran\c{c}ois Pomerleau%
	\thanks{The authors are from the Northern Robotics Laboratory, Universit\'{e} Laval, Canada.	\newline
		{\tt\small philippe.babin.1@ulaval.ca }}}
\maketitle
\abstract{
The ability to map challenging subarctic environments opens new horizons for robotic deployments in industries such as forestry, surveillance, and open-pit mining.
In this paper, we explore possibilities of large-scale lidar mapping in a boreal forest.
Computational and sensory requirements with regards to contemporary hardware are considered as well.
The lidar mapping is often based on the \ac{SLAM} technique relying on pose graph optimization, which fuses the \ac{ICP} algorithm, \ac{GNSS} positioning, and \ac{IMU} measurements. 
To handle those sensors directly within the \ac{ICP} minimization process, we propose an alternative approach of embedding external constraints.
Furthermore, a novel formulation of a cost function is presented and cast into the problem of handling uncertainties from \ac{GNSS} and lidar points.
To test our approach, we acquired a large-scale dataset in the \textit{For\^{e}t Montmorency} research forest.
We report on the technical problems faced during our winter deployments aiming at building 3D maps using our new cost function.
Those maps demonstrate both global and local consistency over \SI{4.1}{\km}.
}
\keywords{Lidar, cost function, ICP, penalties, mapping, GNSS, GPS, winter, forest, anisotropic}

\section{Introduction}
\label{sec:intro}
\acresetall

Autonomous mobile robots require a representation (i.e., a map) of the environment in order to perform specific tasks.
For instance, maps are needed internally to plan motions and avoid obstacles.
The map itself can also be the objective, captured by robots and used for later analysis \cite{Zlot2014b}, including forestry inventories~\cite{Pierzchala2018}.
Although many solutions exist for localization and mapping, the environment itself influences the complexity of the task, and thus dictates which algorithm to use.
In this study, we targeted snowy, subarctic forests to explore new challenges to large-scale mapping. 
Indeed, this type of environment is minimally-structured, making registration more difficult.
Moreover, the ruggedness of terrains dictates the need for a full six \ac{DoF} solution, with little assumption on trajectory smoothness.

\begin{SCfigure}
	\centering
	\includegraphics[width=0.72\textwidth,trim={0mm 7mm 0mm 5mm}, clip]{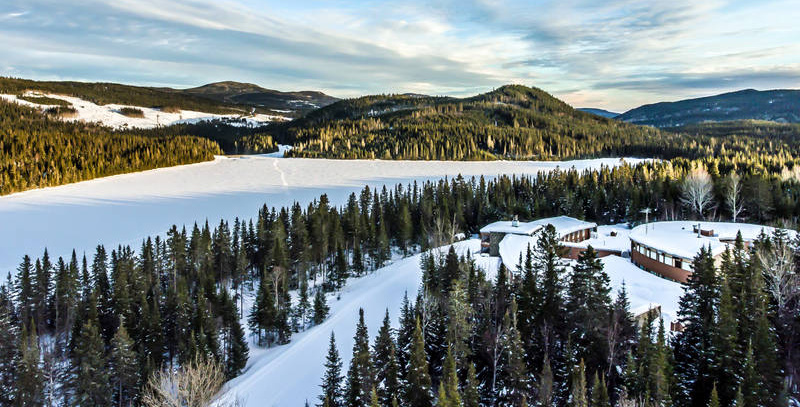}
	\caption{
		The boreal forest \textit{For\^{e}t Montmorency} presents an environment ideal to test robotic applications in a subarctic region. For lidar mapping, it offers challenging dense forest conditions. Credit: \textit{For\^{e}t Montmorency}.
	}
	\label{fig:montmorency_picture}
\end{SCfigure}

Another difficulty brought by subarctic environments is the lack of distinctive visual features during snowy periods~\cite{Williams2012}.
With images of snowy surfaces, it is challenging to extract enough features in order to support visual odometry or vision-based \ac{SLAM}.
This precludes the use of passive camera-based localization systems, making lidar the sensing modality of choice for these conditions.
A natural approach to mapping in this case is to incrementally build a 3D point cloud map from scans taken at different locations, using the \ac{ICP} algorithm~\cite{Besl1992}.
The \ac{ICP} algorithm iteratively finds corresponding points between two point clouds and looks for a rigid transformation minimizing an alignment error.
This approach ensures local consistency, yet inevitably suffers from global drift \cite{Pomerleau2013}.
This drift problem can be mitigated by \ac{SLAM} techniques based on pose graph optimization \citep{Thrun2006,Cadena2016}. 
The key idea behind the latter is to identify loop closures and optimize all estimated transformations between the individual lidar scans to ensure global consistency.
Unfortunately, in environments which do not allow loop closure (such as a long straight trajectory), drift cannot be avoided without additional external localization sources, such as \ac{GNSS}. 
Moreover, the pose graph optimizer requires uncertainty estimations in the form of covariance matrices for all transformations, including those from \ac{ICP}.
As shown by \citet{Landry2019}, this uncertainty can be modeled, learned or sampled, but the Gaussian distribution assumption does not hold up well in complex 3D environments.

On the other hand, autonomous robots operating on polar ice sheets \cite{Lever2013} often rely on \ac{GNSS} as their main source of positioning.
In unstructured environments (e.g., boreal forests, taiga), \ac{GNSS} cannot be used this way due to high uncertainty of position estimates.
This uncertainty is caused by interference of the canopy with the signals from satellites~\citep{Jagbrant2015}.
Still, the main advantage of \ac{GNSS} is that it provides a global source of positioning, which shows minimal and bounded bias compared to the \ac{ICP}.
Meanwhile, \ac{ICP} creates maps that are crisp (i.e. locally consistent).

The goal of our paper is to demonstrate large-scale mapping of difficult environments, while generating maps that are \emph{i)} crisp, \emph{ii)} without long-term drifts, and \emph{iii)} that can be updated swiftly.
To satisfy the first two criteria, we experimented with embedding external constraints directly within the \ac{ICP} minimization process using a novel formulation to handle uncertainty on positions.
More precisely, we propose to augment the \ac{ICP} algorithm by adding penalty terms based on the global \ac{GNSS} positioning and the \ac{IMU} attitude estimates, weighed by their uncertainty.
This formulation has the advantage that the uncertainty associated with these external constraints, contrary to the \ac{ICP}'s, can usually be readily estimated.
Second, it avoids undue oscillations induced by alternating between graph minimization and point cloud registration, as both algorithms have no guarantees of sharing the same minimum.
The third criterion pertains to the need for fast point-cloud map update in autonomous systems.
The main bottleneck of the \ac{ICP} algorithm is the update of the KD-tree, a data structure used for a fast nearest-neighbor search.
Since our objective is to build large maps, this slowdown becomes unacceptable. %
We investigate a simple optimization technique to reduce the execution time of registrations by reducing the size of the map portion used for the KD-tree update.
We tested our approach on large-scale maps recorded in a boreal forest during the winter (shown in \autoref{fig:montmorency_picture}). A particular emphasis has been placed on discussing problems related to the different aspects of lidar mapping for this type of environment.

\section{Related Work}
\label{sec:relatedwork}

The context our work involves robotic deployments in harsh, snowy environments.
This problem of mapping and localization in these conditions has only been investigated by a few publications.
For example, visual \ac{SLAM} for a robotic snowmobile platform was deployed by \cite{Williams2009} on a glacier in Alaska.
The authors report difficulties relating to the relatively low number of visual features in close vicinity of the mapping platform, compared to visual features located on the horizon.
Since the nearby features are vital for translation estimation, image processing techniques are proposed to improve their extraction.
Effects of changing shape and appearance of snowy areas on a path-following algorithm are also discussed in \cite{Paton2016a}.
Their findings further motivate the use of lidars for mapping in these conditions.
In our approach, the deployment of the lidar sensor translates the problem of extracting image features for localization into the problem of locating against 3D geometry.
Areas covered by the boreal forests comply well with this requirement. On the other hand, on open plateaus where 3D features are sparse, the \ac{GNSS} constraints assure consistent localization and mapping.
Moreover, similarly to \cite{Williams2009} and \cite{Paton2016a}, we do not require wheel or track odometry measurements from the mobile platform.
This feature simplifies integration of the mapping system into different mobile vehicles which do not offer any odometry (in our case, the sleigh).
We, however, benefit from an \ac{IMU} which provides attitude prior with unbiased roll and pitch angles.

Laser scans can be captured from a ground-based static sensor as in the work of \cite{McDaniel2012} who employed a stationary high density lidar sensor for terrain classification and identification of tree stems.
Alternatively, the \ac{ALS} approach allows the mapping of vast forested areas from the air.
Besides the \ac{ALS}, \emph{Structure from Motion} technique \cite{Wallace2016} and stereo imagery \cite{Tian2013} are further alternatives to creating 3D maps, suitable mainly for light aerial drones.
Our goal is creating and maintaining globally consistent 3D maps for autonomous ground robots.
In the case of \ac{ALS}, the global consistency is easier to achieve because of the high-altitude point of view and unobstructed \ac{GNSS} reception.
Contrarily, ground robots only observe a limited portion of the area at a time and their \ac{GNSS} reception is partially occluded by the canopy.
On the other hand, ground-based 3D scans offer high details, also useful for in-depth vegetation analysis.
The problem of storing and managing large amounts of data is common to all of the mentioned works.
In our approach, we propose a technique to limit the computation demands during the mapping process. 

Fueled by the increasing interest in self-driving cars, multiple large-scale urban datasets, most notably \citep{Geiger2013IJRR}, (containing lidar and \ac{GNSS} information beside other sensors) have become available.
These datasets have accelerated development, refining a variety of visual- and lidar-based \ac{SLAM} algorithms.
Contrary to structured urban environments, we investigate the characteristics of mapping in unstructured ones (forests) in harsh winter conditions.
Additionally, any improvement on the accuracy of registration algorithms reduces pressure on loop-closure and graph minimization algorithms, leading to more robust lidar-based \ac{SLAM} algorithms overall.

From the extensive family of \ac{ICP} variants \cite{Pomerleau2015b}, our contribution relates mainly to incorporating generalized noise models into the \ac{ICP} algorithm.
Since the \ac{GNSS} positioning provides a confidence estimate in the form of a covariance matrix, simplification to an isotropic noise model discards potentially important information.
\citet{Ohta1998} were first to consider anisotropic and inhomogeneous noise models when estimating optimal rotation of features extracted from stereo-pair depth images for 3D reconstruction.
Later, the \ac{GTLS-ICP} algorithm was introduced by \citet{Estepar2004} for registering medical fiduciary markers.
The work considers an anisotropic noise model in the registration phase of \ac{ICP} and accounts for optimizing translation component as well.
Further improvements were introduced by \citet{Maier-hein2012}, where the matching phase is modified to benefit both from KD-tree search speed and Mahalanobis-distance metric.
This technique has been eventually enhanced by the introduction of a new kind of KD-tree which directly supports Mahalanobis-distance and a new minimizer~\cite{Billings2015}.
In our approach, the anisotropic noise is strictly limited to the \ac{GNSS} position measurements, making the problem slightly different. %
We look for a way to integrate this positioning information together with its anisotropic noise into the \ac{ICP}.
More closely related to robotic applications, the \ac{GICP} algorithm
\cite{Segal2009} preserves the possibility to model measurement noise, while focusing on minimizing the plane-to-plane metric.
However, using an iterative minimizer such as \ac{BFGS} inside the matching loop of \ac{ICP} is prohibitively slow.
In this paper, we investigate how to link point set registration to include penalty terms brought by the \ac{GNSS} and \ac{IMU} measurements within the same mathematical framework generic to anisotropic noise.

\section{Theory}
\label{sec:theory}

\subsection{New Formulation for Point-to-Gaussian Cost Function}
\label{subsec:p_to_gauss}
The \ac{ICP} algorithm aims at estimating a rigid transformation $\widehat{\bm{T}}$ that best aligns a set of 3D points $\mathcal{Q}$ (i.e., a \emph{map} point cloud)  with a second set of 3D points $\mathcal{P}$ (i.e., \emph{scan} point cloud), given a prior transformation $\widecheck{\bm{T}}$. 
For a better representation of surfaces, the points of the map point cloud can be represented locally by planes.
The problem of rigid registration using points from the scan and planes from the map \citep{Chen1992} can be summarized as minimizing the point-to-plane cost function $\mathrm{J}_\text{p-n}(\cdot)$ following
\begin{align}
\label{eq:icp_p_to_plan_min}
\widehat{\bm{T}} &= \arg \min_{\bm{T}} \: {J}_\text{p-n} \left( \mathcal{Q}, \mathcal{P}, \widecheck{\bm{T}} \right)  
\text{, \hspace{0.2cm} with}
\\
{J}_\text{p-n} &= \sum_{i=1} \sum_{j=1} w_{ij} (\bm{e}_{ij}^{T} \bm{n}_i)^2
\text{, \hspace{0.2cm} and \hspace{0.2cm}}
\bm{e}_{ij} = \bm{q}_i - \widecheck{\bm{R}} \bm{p}_j - \widecheck{\bm{t}},
\end{align}
where $\bm{e}_{ij}$ is the error vector between the $i$\textsuperscript{th} point $\bm{q}$ of $\mathcal{Q}$ and the $j$\textsuperscript{th} point $\bm{p}$ of $\mathcal{P}$, $\bm{n}_i$ is the normal of the plane, $\widecheck{\bm{R}}$ and $\widecheck{\bm{t}}$ are respectively the rotation and translation part of $\widecheck{\bm{T}}$, and $w_{ij}$ is a weight limiting the impact of outliers as surveyed by \citet{Babin2019}.
The double summation in (\ref{eq:icp_p_to_plan_min}) is expensive to compute and is typically approximated using a subset of pairs using nearest neighbor points of each scan point.
To simplify the notation, we will use $\bm{e}_m$ for each error to be minimized, with $m$ being the index of this subset.
Point-to-plane error outperforms point-to-point error in most cases~\citep{Pomerleau2013}.
However, it does not represent non-planar surface well.
Point-to-Gaussian provides a more versatile representation \citep{Balachandran2009}. 
Instead of being represented by a plane, each point in $\mathcal{Q}$ is the mean of a Gaussian and its incertitude is represented by a covariance.
The point-to-Gaussian cost function $\mathrm{J}_\text{p-g}$ thus becomes the following:
\begin{equation}
\label{eq:icp_cov_min}
\mathrm{J}_\text{p-g} =  \sum_{m=1} \left(\bm{e}^T \bm{W}^{-1} \bm{e}\right)_{m},
\end{equation}
where $\bm{W}^{-1}$ is the inverse of the covariance.
In point-to-Gaussian, the Mahalanobis distance is minimized instead of the Euclidean distance (point-to-point) or the projected distance to a plan (point-to-plane).
Instead of using a second iterative solver within the matching loop of \ac{ICP} \citep{Estepar2004, Segal2009, Billings2015}, we propose a novel decomposition to minimize the point-to-Gaussian error (\ref{eq:icp_cov_min}) directly using the  equations for point-to-plane error (\ref{eq:icp_p_to_plan_min}).
The inverse of the covariance $\bm{W}^{-1}$ can be expressed as a matrix $\bm{N}$ of eigenvectors and a diagonal matrix $\bm{\Lambda}$ holding the sorted eigenvalues, with $\lambda_1 < \lambda_2 < \lambda_3$, using
\begin{equation*}
\bm{W} = \bm{N} \bm{\Lambda} \bm{N}^{T} \Rightarrow \bm{W}^{-1} =  \bm{N} \bm{\Lambda}^{-1} \bm{N}^{T} .
\end{equation*}
The decomposition can be inserted inside the cost function and reformulated as three point-to-plane errors using a projection for each of the eigenvector.
Dropping the summation and the indices for clarity, we obtain for a single pair of points
\begin{align}
\mathrm{J}_\text{p-g} &=  \bm{e}^T \bm{N} \bm{\Lambda}^{-1} \bm{N}^{T} \bm{e} 
\\
&=  \bm{e}^T 
\bbm \bm{n}_1 & \bm{n}_2 & \bm{n}_3 \ebm 
\mathrm{diag}\left(\tfrac{1}{\lambda_1}, \tfrac{1}{\lambda_2}, \tfrac{1}{\lambda_3}\right)
\bbm \bm{n}_1 & \bm{n}_2 & \bm{n}_3 \ebm^T
\bm{e}
\nonumber \\
&= 
\underbrace{\frac{1}{\lambda_1} \Big( \bm{e}^T \bm{n}_1\Big)^2}_{\mathrm{J}_\text{p-n}} +
\frac{1}{\lambda_2} \Big( \bm{e}^T \bm{n}_2\Big)^2 +
\frac{1}{\lambda_3} \Big( \bm{e}^T \bm{n}_3\Big)^2 
,
\label{eq:pt-g_3D}
\end{align} 
where $\lambda_i$ is an eigenvalue and $\bm{n}_i$ is its associated eigenvector.
Thus, point-to-Gaussian can be used with any point-to-plane minimizer.
In fact, point-to-plane is a special case of point-to-Gaussian, where the first eigenvalue $\lambda_1$ is small enough compared to $\lambda_2$ and $\lambda_3$.
This formulation can be used to also minimize Gaussian-to-Gaussian by setting $\bm{W}$ to the sum of the uncertainty of point $\bm{q}$ with the rotated uncertainty of its associated $\bm{p}$.

\subsection{Adding Penalty Terms to \ac{ICP}}
\label{subsec:theo_penalty}
\ac{ICP} mapping creates crisp maps by taking into account local geometric characteristics contained in each new point cloud. 
Therefore, global consistency is not enforced.
On the other hand, \ac{GNSS} provides globally consistent positioning, but yields low local precision, especially when compared to \ac{ICP} in forested areas.
Furthermore, there is a disproportion between the altitude, latitude and longitude positioning components, the altitude being the least precise.
By fusing \ac{ICP}, \ac{GNSS} and \ac{IMU} information, we propose to compensate for the \ac{ICP} drift. %

Penalties are a natural way to add a constraint to the minimization step of \ac{ICP}.
They can be seen as imaginary points added to the point cloud during minimization for which the association is known. The minimization problem thus becomes:
\begin{equation}
\label{eq:icp_min_with_penalties}
\widehat{\bm{T}} = \arg \min_{\bm{T}}  
\underbrace{\frac{1}{M} \sum_{m=1} \left(w \bm{e}^T \bm{W}^{-1} \bm{e} \right)_{m}}_{\text{point clouds}}  
+ \underbrace{\frac{1}{K} \sum_{k=1} \left( \bm{e}^T \bm{W}^{-1} \bm{e} \right)_{k}}_{\text{penalties}},
\end{equation}
where $\bm{e}_{k}$ and $\bm{W}_k$ are respectively the error and the covariance of the $k$\textsuperscript{th} penalty, $M$ is the number of matched points and $K$ the number of penalty added.
The penalty error $\bm{e}_{k}$ consists of two points added to their respecting frames of reference
\begin{equation}
\label{eq:penalties_error}
\bm{e}_k = \prescript{\text{M}}{}{\bm{q}_k} - \prescript{\text{M}}{S}{\widecheck{\bm{T}}} \prescript{\text{S}}{}{\bm{p}_k},
\end{equation}
where $\prescript{\text{M}}{}{\bm{q}_k}$ is the position of the $k$\textsuperscript{th} penalty in the map frame, $\prescript{\text{M}}{S}{\bm{T}}$ is the transformation from the scan frame (S) to the map frame (M) at the current iteration and $\prescript{\text{S}}{}{\bm{p}_k}$ is the position of the  $k$\textsuperscript{th} penalty in the scan frame.
For instance for the \ac{GNSS} penalty, $\prescript{\text{M}}{}{\bm{q}_k}$ is the \ac{GNSS}'s global position and $\prescript{\text{S}}{}{\bm{p}_k}$ is the origin of the scan frame.
Effects of adding penalty points are presented in \autoref{sec:res_penalty}.

Since the \ac{GNSS} penalty points come in the form of a Gaussian distribution and point clouds contain plane normal information, equation (\ref{eq:icp_min_with_penalties}) is approximated using a constant scale $s$ in our implementation. 
This constant ensures the conversion from projected distances to Mahalanobis distances by assuming that $\frac{1}{\lambda_1}$ is constant for all points and neglecting  $\frac{1}{\lambda_2}$ and$\frac{1}{\lambda_3}$ in (\ref{eq:pt-g_3D}).
In this setting, the final cost function to be optimized becomes:
\begin{equation}
\label{eq:real_icp_min_with_penalties}
\mathrm{J}_\text{p-g} \approx  
\frac{s}{M} \sum_{m=1} \Big(w (\bm{e}^{T}  \bm{n}_{1})^2 \Big)_{m}
+ 
\frac{1}{K} \sum_{k=1} \Big( \bm{e}^T \bm{W}^{-1} \bm{e}\Big)_{k}.
\end{equation}
\subsection{Iterative Closest Point mapping}
\label{subsec:mapping}

The mapping was achieved using a modified version of \texttt{ethz-icp-mapping}~\citep{Pomerleau2014}.
The mapper performs the following steps: 1) Move the scan to the initial estimate, 2) register the scan with the map using \ac{ICP}, and 3) insert the scan inside the map.
The initial estimate $\widecheck{\bm{T}}$ is composed of a translation increment based on the \ac{GNSS} positioning and change in orientation based on the \ac{IMU}.
The \ac{IMU} heading is corrected by the \ac{GNSS} positioning as long as the platform moves forward.
Justification of this correction and a possible alternative are discussed in \autoref{sec:res_penalty}.
Since this initial estimate $\widecheck{\bm{T}}$ is utilized in an incremental manner, the mapping can diverge over time.
As for the construction of the global map $\mathcal{Q}$, the whole scan $\mathcal{P}$ is not directly concatenated. 
Rather, only points that are farther than $\epsilon$ from any points in $\mathcal{Q}$ are inserted.
This helps in keeping the global map uniform, without sacrificing registration precision.
As the robot explores the environment, the complexity of registration grows linearly with the number of points in the map due to the KD-tree structure updates.
To stabilize the mapping complexity, a scan $\mathcal{P}$ is not registered against the whole map $\mathcal{Q}$, but only against a subsection of the map within a radius $r_\text{max}$ equal to the maximum range of the lidar.
The effects of this optimization is shown in \autoref{sec:res_realtime}.

\section{Experimental Setup}
\label{sec:exp_setup}
\subsection{Data Acquisition Platform}
\begin{SCfigure}
	\centering
	\includegraphics[trim={2cm 0 0 4.8cm}, clip, width=0.4985\textwidth]{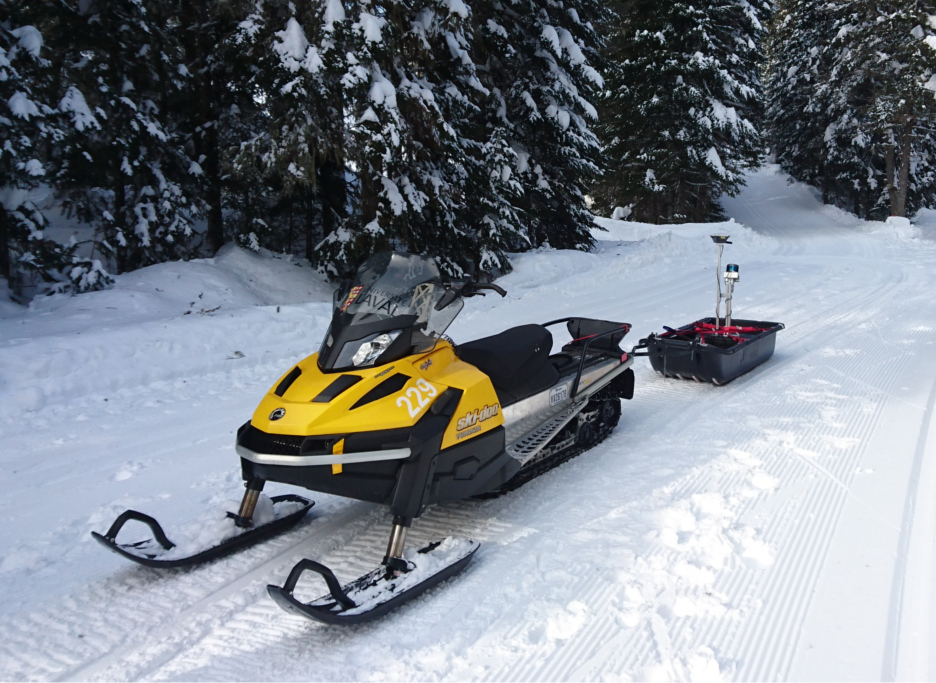}
	\includegraphics[width=0.216\textwidth]{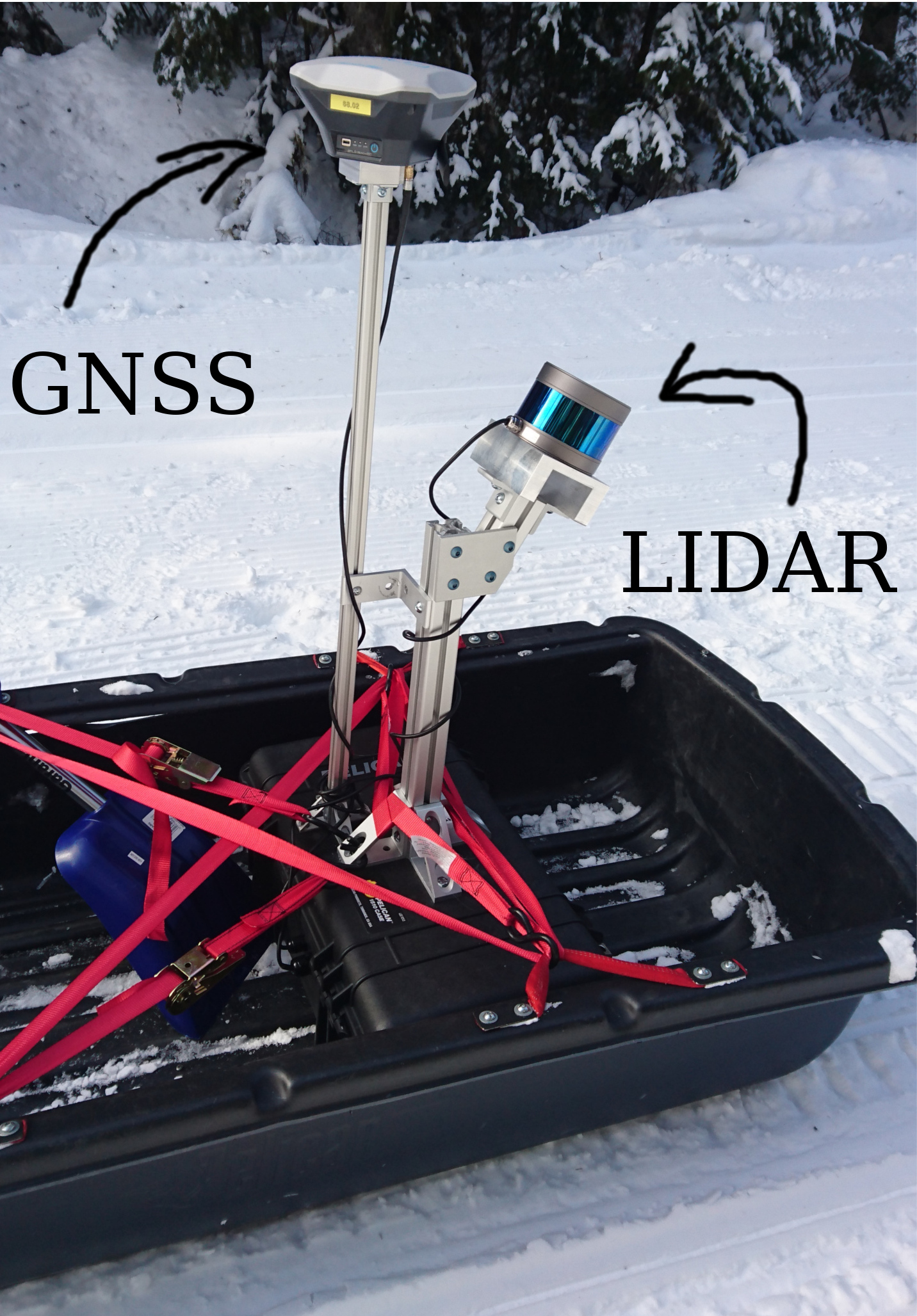}
	\caption{
		The data acquisition platform mounted on a sleigh behind the snowmobile (\emph{left}).
		This configuration limited transfer of engine vibrations to the sensors.
		\emph{Right}: close-up of the \ac{GNSS} receiver and the RS-16 lidar.
	}
	\label{fig:photos-platform}
\end{SCfigure}

For our experiments, we developed a rugged data acquisition platform which can withstand snow and sub-zero temperatures.
It comprises an Xsens MTI-30 \ac{IMU}, a Robosense RS-16 lidar, and a REACH RS+ GNSS antenna powered by two 20Ah 12V AGM batteries (\SI{10}{\hour} battery life).
A small, low-power computer (AIV-APL1V1FL-PT1) records the sensor data using the \ac{ROS} framework.
This platform can be attached to most of mobile vehicles (see \autoref{fig:photos-platform}).
The rotation axis of the lidar sensor is at an angle of \ang{27} from the vertical.
This orientation has been chosen for two reasons: 1) the lidar does not see the mobile vehicle nor its operators as long as they are in front of the platform, and 2) as mentioned in \citep{Laconte2019} a lower incidence angle with the ground reduces the odometry drift.
The \ac{GNSS} antenna is coupled with a fixed station (also a REACH RS+ antenna) mounted on a tripod to provide a \ac{RTK} solution.

\begin{SCfigure}
	\centering
	\includegraphics[width=0.6\textwidth]{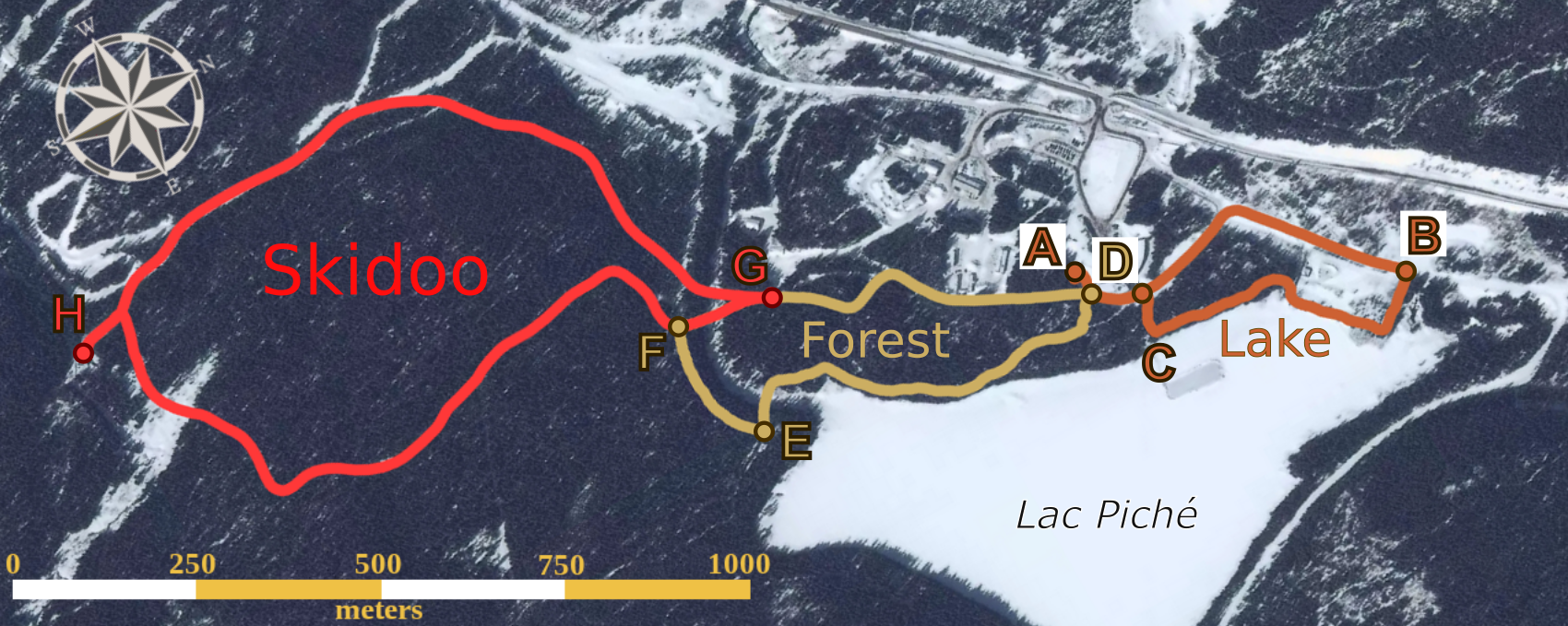}
	\caption{
		The three large loops completed in the Montmorency forest in order to collect the \texttt{lake} (\SI{0.8}{\kilo\meter}), \texttt{forest} (\SI{1.3}{\kilo\meter}) and \texttt{skidoo} (\SI{2}{\kilo\meter}) datasets.
		The map was adapted from \textcopyright Mapbox  and \textcopyright OpenStreetMap contributors.
	}
	\label{fig:satelite_map}
\end{SCfigure}

The Universit\'{e} Laval owns the \textit{For\^{e}t Montmorency}, the largest research forest in the world with over \SI{412}{\kilo\meter\squared} of boreal forest (\autoref{fig:montmorency_picture}).
For our experiments, we collected data along three large loops (see \autoref{fig:satelite_map}).
Two of them (i.e., \texttt{lake} and \texttt{forest}) consist of a mix of narrow walkable trails and wider snowmobile trails, while the last one (i.e., \texttt{skidoo}) followed exclusively a wide snowmobile trail.

More specifically, the dataset \texttt{lake} was recorded by mounting our platform to a snowmobile (see \autoref{fig:photos-platform}) and then having two operators pull the sleigh through a pedestrian trail.
The snowmobile drove through a cross-country skiing trail, which is an open area with good \ac{GNSS} coverage (\texttt{A} to \texttt{B} in \autoref{fig:satelite_map}).
The pedestrian trail was a dense forest path (\texttt{B} to \texttt{C}), where the platform suffered from poor \ac{GNSS} reception due to the tree canopy.
The overlap between scans diminished abruptly each time branches came near the sensor.
Similarly, the dataset \texttt{forest} consists of a pedestrian trail and a cross-country skiing trail.
In the first part of the trajectory, a pedestrian trail in the dense forest was traversed with a sleigh (\texttt{D} to \texttt{E}), followed by an untapped path through an even denser forest (\texttt{E} to \texttt{F}) and finally the sleigh was attached to a snowmobile and driven back to the starting point (\texttt{F} to \texttt{D}).
Lastly, the dataset \texttt{skidoo} follows a \SI{2}{\kilo\meter} long snowmobile trail.
The data were gathered during a light snow fall.
This loop was the easiest to map because it provided clear \ac{GNSS} signal and was relatively flat from beginning to end (\texttt{G} to \texttt{H} to \texttt{G}).

\section{Field Results}
\label{sec:results}

\vspace{-5pt}
\subsection{Effects of Adding \ac{GNSS} Penalty to \ac{ICP}}
\vspace{-5pt}
\label{sec:res_penalty}

In order to inject the \ac{GNSS} positioning information as a constraint into the \ac{ICP} algorithm, it is necessary to find the correct transformation between the \ac{ICP} map coordinate frame and the local tangent \ac{ENU} frame.
The translation, roll and pitch angles are directly observable from the \ac{GNSS} receiver and the \ac{IMU}.
The yaw angle (heading) measurement provided by the magnetometer part of the \ac{IMU} is, however, affected by soft- and hard-iron errors induced by the mobile platform itself and by local deviations of the Earth's magnetic field.
A practical solution to this problem was to observe a short initial portion of the \ac{GNSS} trajectory to estimate the magnetometer heading offset.
This way, the typical error of between \SI{15}{\degree} and \SI{20}{\degree} could be reduced to \SI{3}{\degree} or less, which led to a satisfactory initial alignment.
Another approach would be attaching a second \ac{GNSS} receiver antenna to the mobile platform and estimating the heading angle from the relative positions of the two antennas.
Both approaches, however, require a precise \ac{RTK} \ac{GNSS} solution.
The standard uncorrected \ac{GNSS} operating under the tree canopy yields excessive error and cannot be used for this purpose.

\begin{SCfigure}
	\centering
	\includegraphics[width=0.70\textwidth]{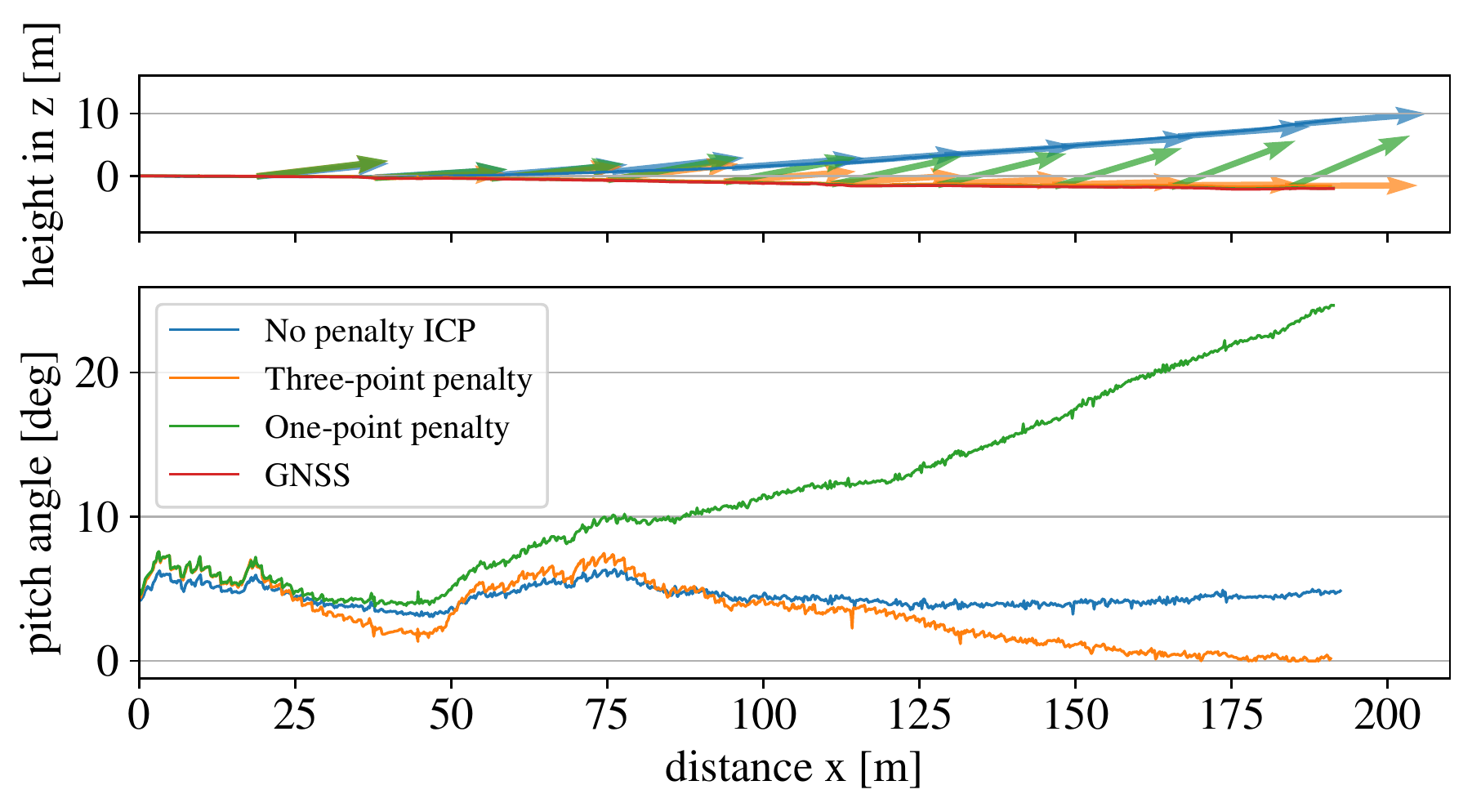}
	\caption{
		Effect of adding penalty points to the \ac{ICP}.
		\emph{Top}: side view of a straight trajectory segment along with orientations represented by arrows.
		\emph{Bottom}: corresponding pitch angles, unavailable for \ac{GNSS}.
		The test was done in an open area with good \ac{GNSS} coverage.
	}
	\label{fig:z_bend}
\end{SCfigure}

We first only applied a single penalty point to the \ac{ICP}, based on the \ac{GNSS} positioning.
As the green trajectory in \autoref{fig:z_bend} demonstrates, this approach does not provide satisfactory results.
On a short straight trajectory, we see that the single-point penalty forces the \ac{ICP} to follow the \ac{GNSS} reference, however, the orientation estimate drifts leading to a malformed map.
In the \autoref{fig:z_bend}, this effect manifests itself as a slow rise in the pitch angle.

To fully constrain the \ac{ICP} and avoid both orientation and position drift, we increased the number of penalty points to three, still following (\ref{eq:real_icp_min_with_penalties}).
The additional two points lie on the gravity and on the heading vectors as follows: In the map frame, one point lies below the \ac{GNSS} position in the direction of the $z$ axis.
The second point lies in the $x$-$y$ plane, in the direction of the current heading as indicated by the \ac{IMU} and \ac{GNSS}.
In the scan frame, these two new points are accordingly projected using the \ac{IMU} orientation information.
In the ideal case, all three penalty points in the scan frame coincide with their map counterparts.
Otherwise, the penalty is forcing the \ac{ICP} solution towards the ideal state.
The effect is demonstrated in the \autoref{fig:z_bend} by the orange trajectory; the \ac{ICP} output follows the \ac{GNSS} positioning while keeping the correct orientation as well.

\vspace{-7pt}
\subsection{The For\^{e}t Montmorency Dataset Results and Discussion}
\vspace{-7pt}
\label{sec:res_datasets}

\begin{figure}[tbp!]
	\vspace{-15pt} %
	\centering
	\copyrightbox[l]{\begin{subfigure}[b]{0.31\textwidth}
		\caption*{Prior map}
		\includegraphics[width=\textwidth]{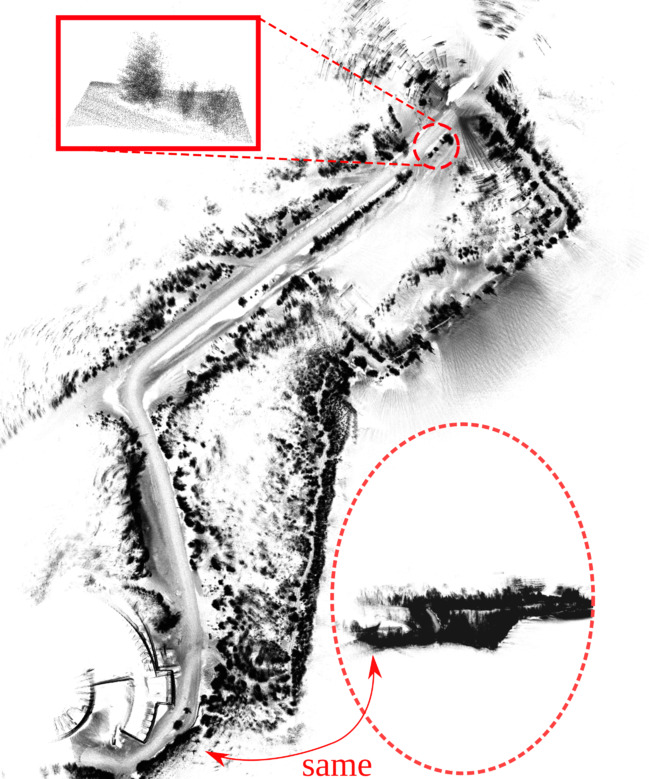}
	\end{subfigure}}{ \textcolor{black}{\hspace{0.7cm}\texttt{lake}}  (\SI{0.8}{\km})}
	~
	\begin{subfigure}[b]{0.3\textwidth}
		\caption*{ICP with penalties}
		\includegraphics[width=\linewidth]{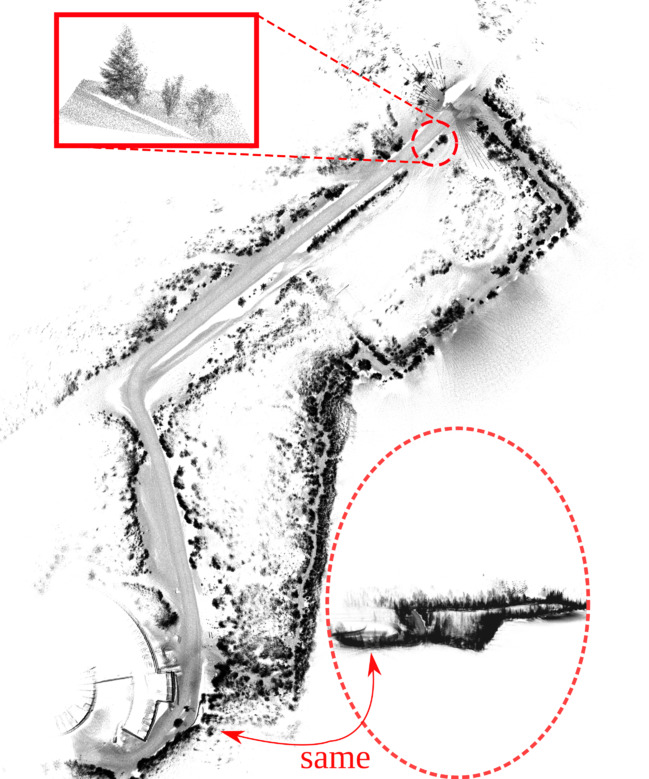}
	\end{subfigure}
	~
	\begin{subfigure}[b]{0.28\textwidth}
		\caption*{ICP without penalty \\(baseline)}
		\includegraphics[width=\linewidth]{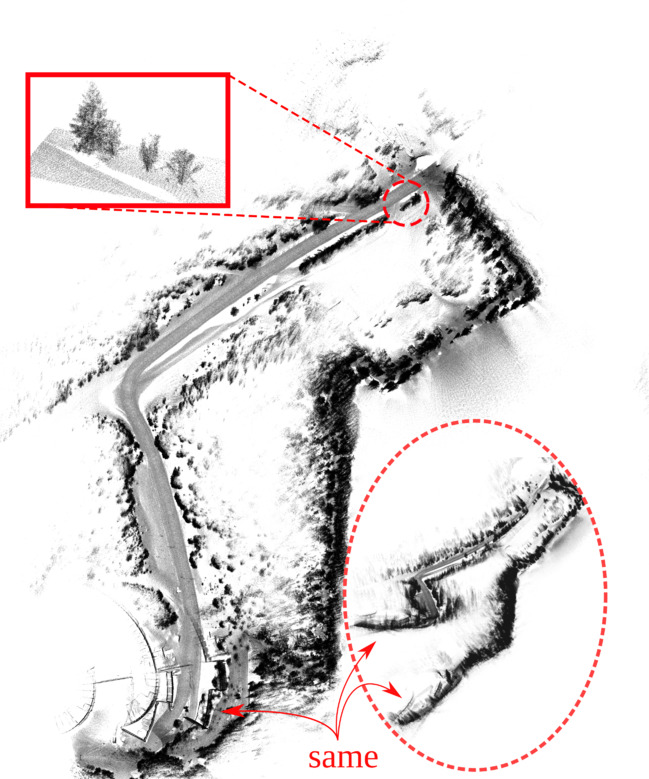}
	\end{subfigure} \vskip -3pt
	\copyrightbox[l]{\begin{subfigure}[b]{0.31\textwidth}
	\includegraphics[width=\textwidth,clip,trim={0cm 0 0 2.8cm}]{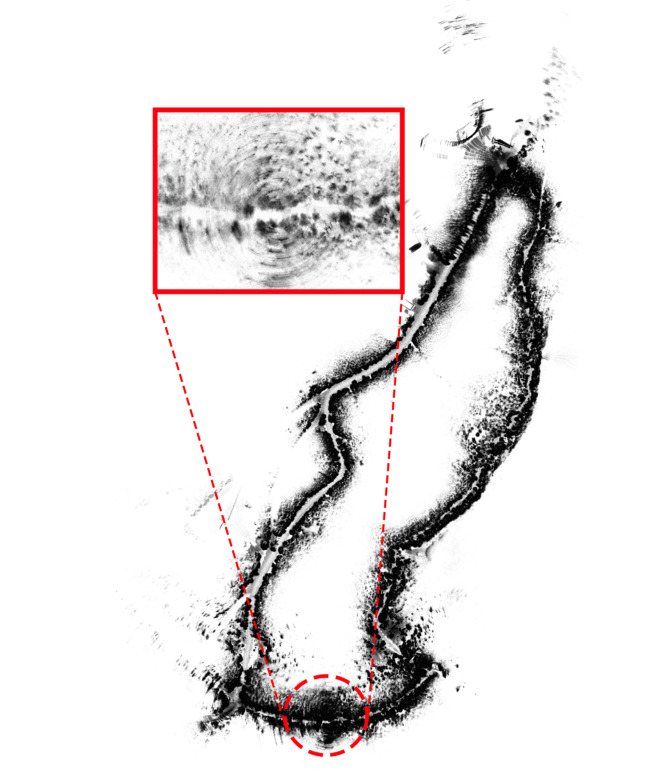}
	\end{subfigure}}{ \textcolor{black}{\hspace{0.5cm}\texttt{forest}} (\SI{1.2}{\km})}
	~
	\begin{subfigure}[b]{0.3\textwidth}
		\includegraphics[width=\linewidth,clip,trim={0cm 0 0 2.8cm}]{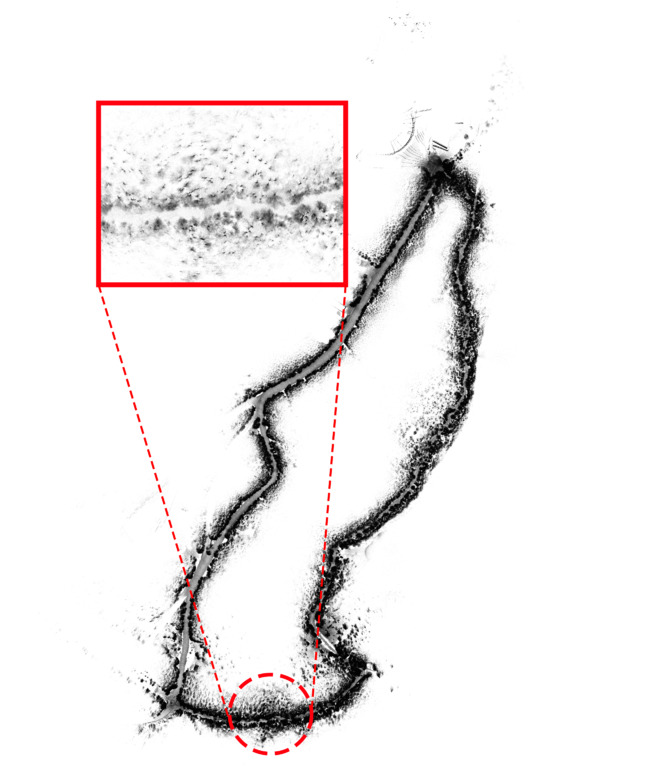}
	\end{subfigure}
	~
	\begin{subfigure}[b]{0.28\textwidth}
		\includegraphics[width=\linewidth,clip,trim={0cm 0 0 1.0cm}]{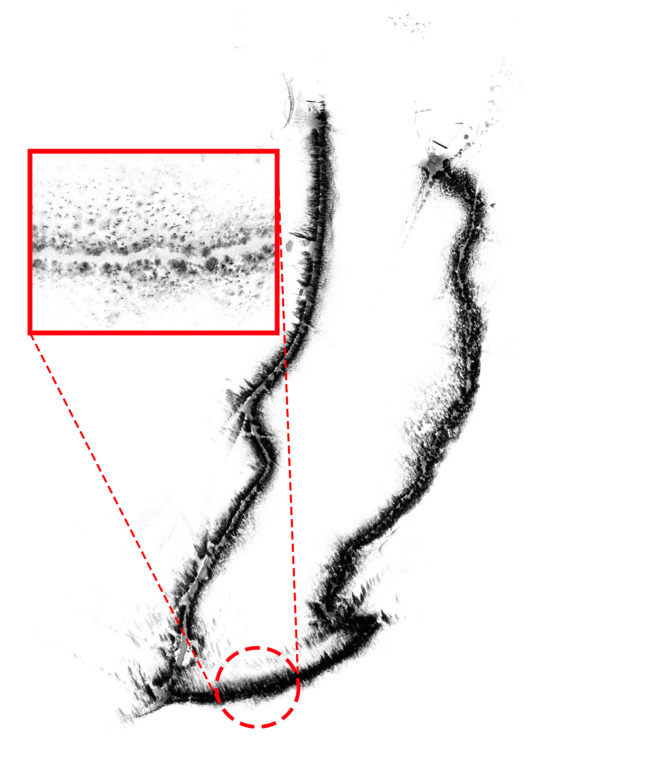}
	\end{subfigure} \vskip -15pt
	\copyrightbox[l]{\begin{subfigure}[b]{0.31\textwidth}
	\includegraphics[width=\textwidth]{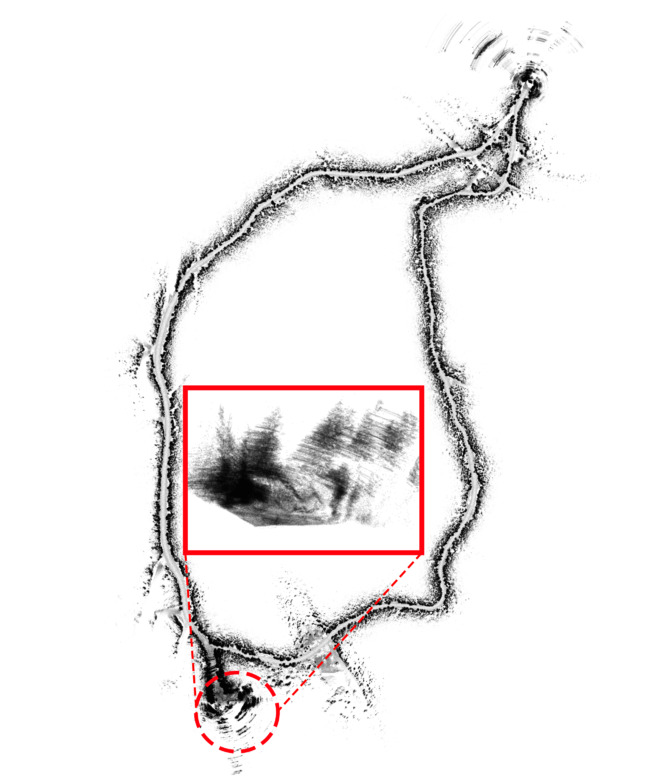}
\end{subfigure}}{ \textcolor{black}{\hspace{0.5cm}\texttt{skidoo}} (\SI{2.0}{\km})}
	~
	\begin{subfigure}[b]{0.3\textwidth}
	\includegraphics[width=\linewidth]{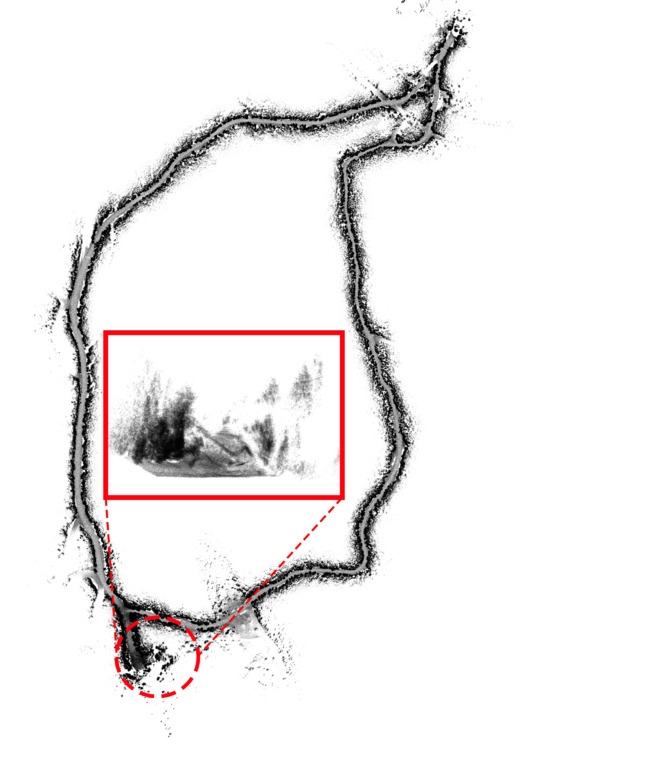}
	\end{subfigure}
	~
	\begin{subfigure}[b]{0.28\textwidth}
	\includegraphics[width=\linewidth]{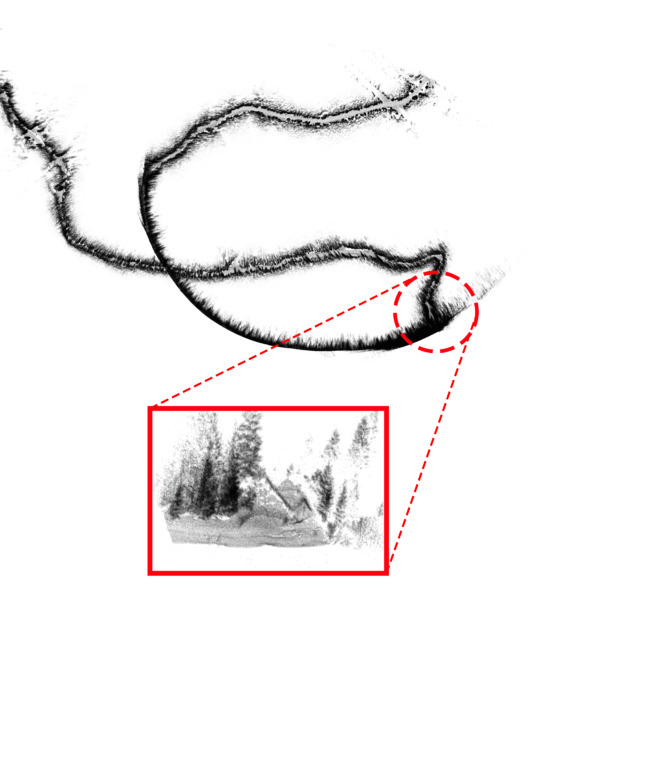}
	\end{subfigure}
	\vspace{-20pt}
	\caption{
		Top view of the point cloud maps of the three datasets; prior, penalty and baseline configurations.
		For the \texttt{lake} dataset, a side view of the map (red dashed ellipse) shows the misalignment between the start and end.
		Red insets show the local (in)consistencies otherwise not visible at the full scale.
		Some trees in the insets are up to \SI{15}{\m} in height.
	}
	\label{fig:prior_vs_penalty_trajectory}
	\vspace{-5pt} %
\end{figure}

For each dataset, three mapping configurations were evaluated: \ac{GNSS}+\ac{IMU} (i.e., prior), \ac{ICP} with penalties (i.e., penalty) and \ac{ICP} without penalty (i.e., baseline). 
When processing, the \texttt{ethz-icp-mapping} was used with $r_{\text{max}} = $ \SI{100}{\meter} and $\epsilon = $ \SI{5}{\centi\meter} for \texttt{lake} and \texttt{forest}.
The \texttt{skidoo} dataset uses $\epsilon = $ \SI{10}{\centi\meter} due to its size and memory requirements.
The resulting maps are shown in \autoref{fig:prior_vs_penalty_trajectory}.
\begin{figure}[tbp!]
	\centering
	\begin{subfigure}[b]{0.245\textwidth}
		\centering %
		\begin{overpic}[width=\linewidth, trim={0 8cm 0 5cm}, clip]{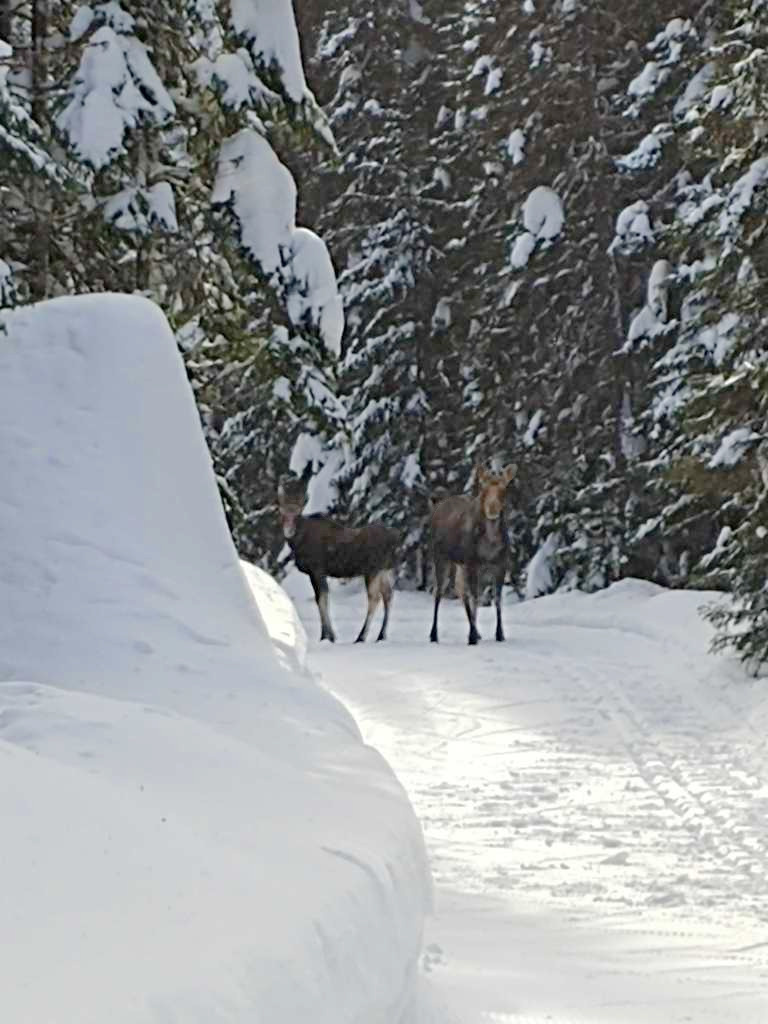}
			\put (5.5, 5.5) {\scriptsize \color{white} \textbf{\textsc{(I) Local Wildlife}}}
			\put (5, 6) {\scriptsize \color{black} \textbf{\textsc{(I) Local Wildlife}}}
		\end{overpic}
	\end{subfigure}
	\begin{subfigure}[b]{0.245\textwidth}
		\centering
		\begin{overpic}[width=\linewidth, trim={2cm 0cm 3cm 0cm}, clip]{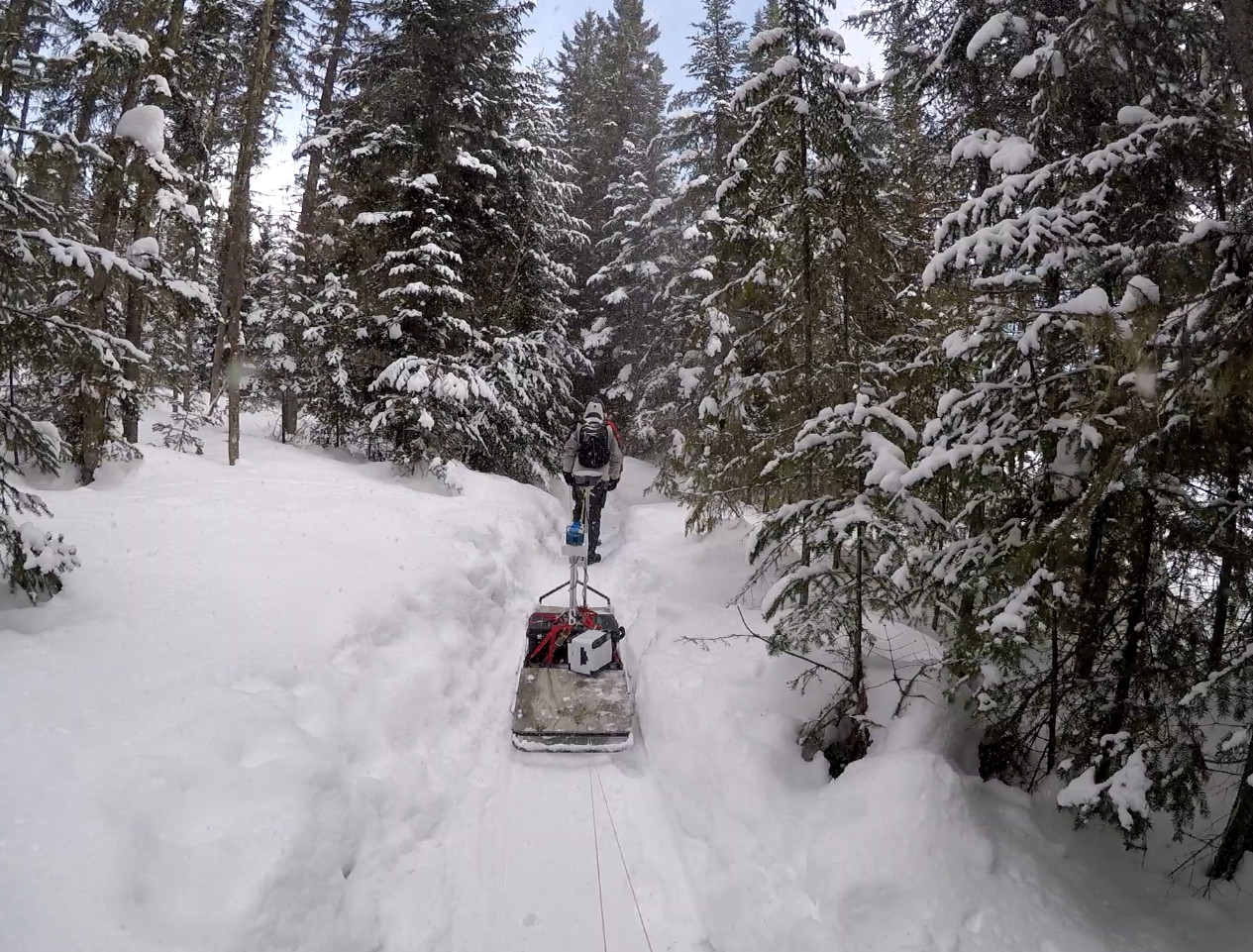}
			\put (5.5, 5.5) {\scriptsize \color{white} \textbf{\textsc{(II) Snow fall}}}
			\put (5, 6) {\scriptsize \color{black} \textbf{\textsc{(II) Snow fall}}}
		\end{overpic}
	\end{subfigure}
	\begin{subfigure}[b]{0.255\textwidth}
		\centering
		\begin{overpic}[width=\linewidth, trim={6cm 11.1cm 6cm 9cm}, clip]{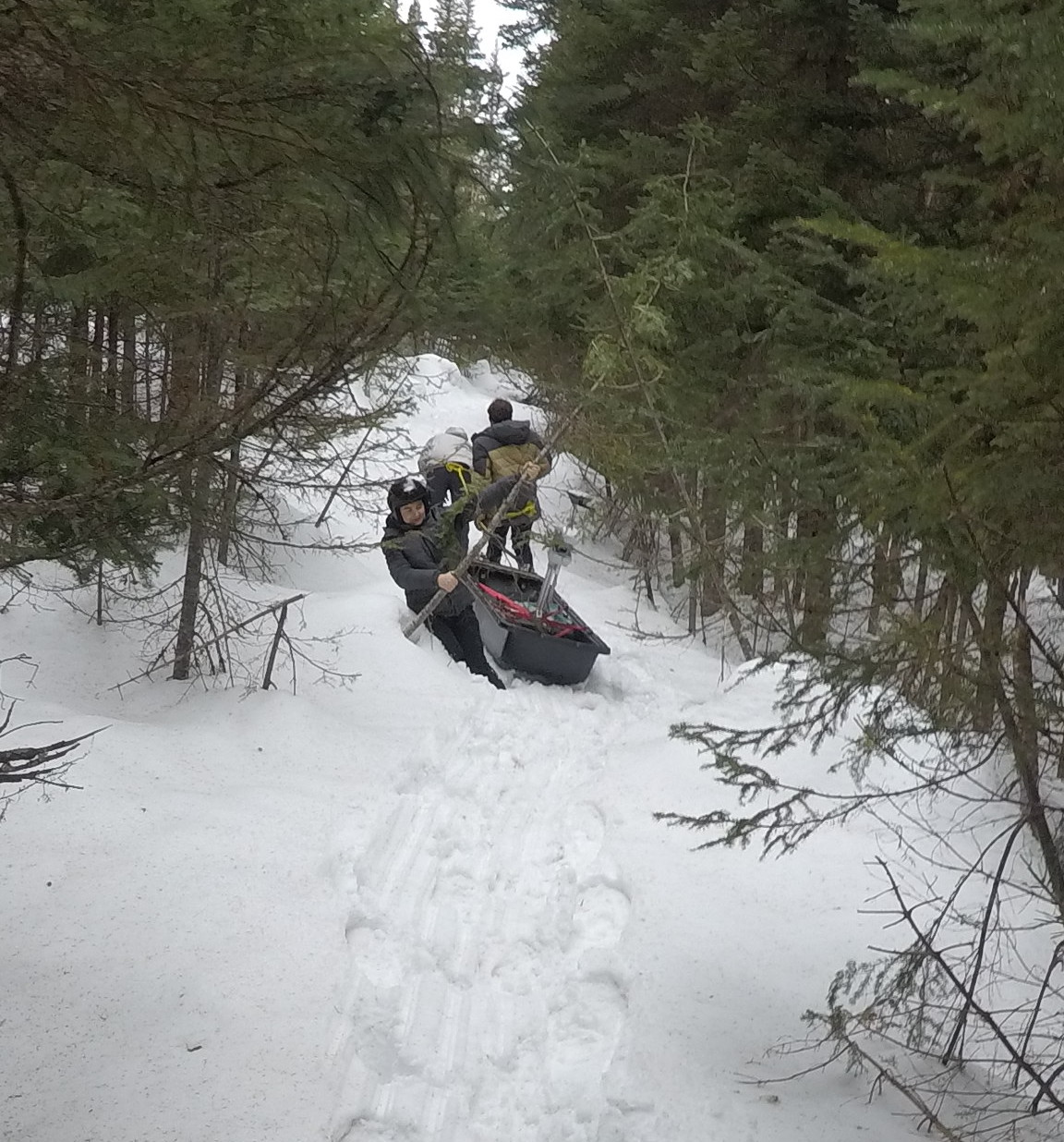}
			\put (5.5,  5.5) {\scriptsize \color{white} \textbf{\textsc{(III) Path Obstacles}}}
			\put (5, 6) {\scriptsize \color{black} \textbf{\textsc{(III) Path Obstacles}}}
		\end{overpic}
	\end{subfigure}
	\begin{subfigure}[b]{0.205\textwidth}
		\centering
		\begin{overpic}[width=\linewidth, trim={1 7cm 1 8.8cm}, clip]{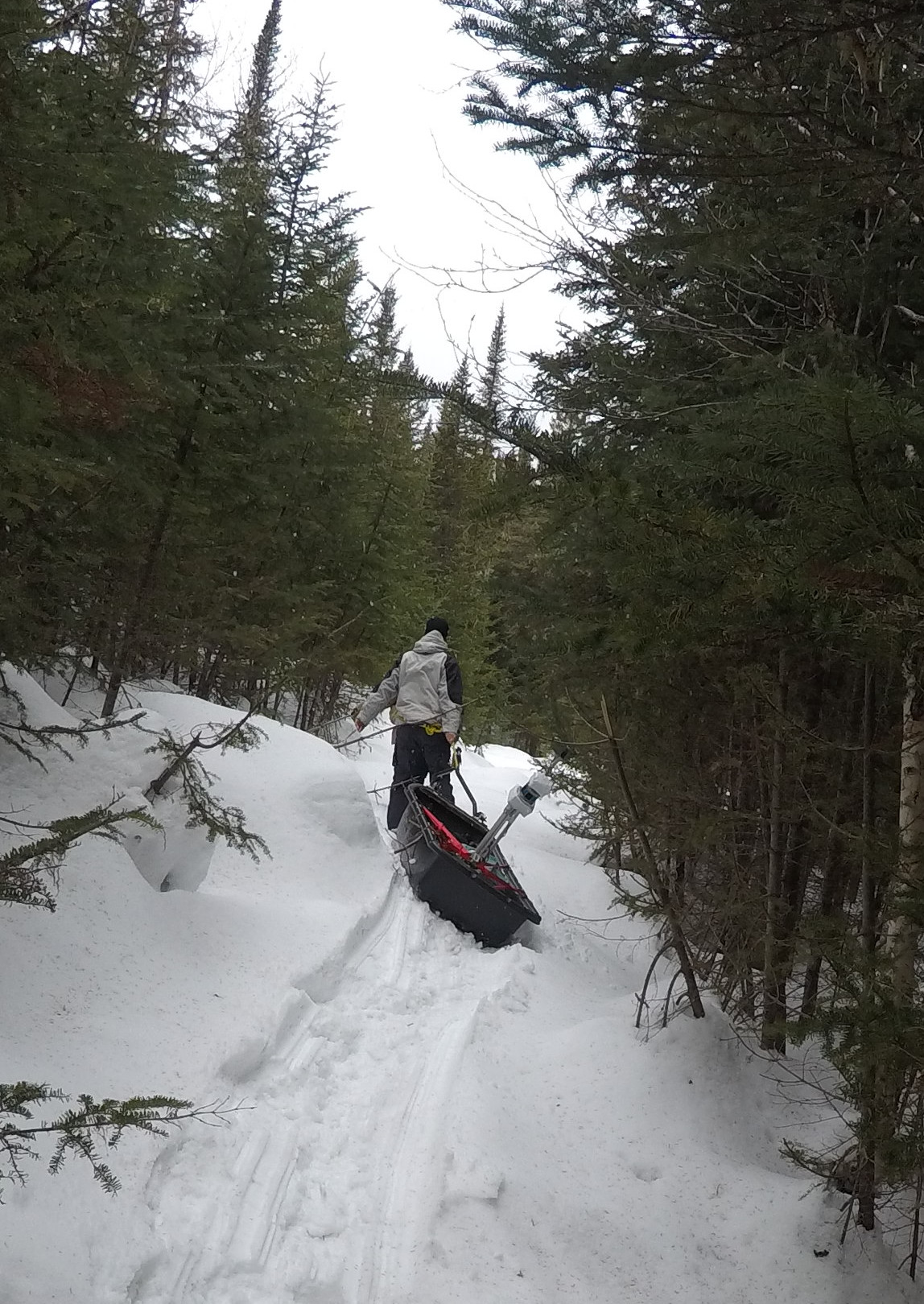}
			\put (5.5,  5.5) {\scriptsize \color{white} \textbf{\textsc{(IV) Uneven Path}}}
			\put (5, 6) {\scriptsize \color{black} \textbf{\textsc{(IV) Uneven Path}}}
		\end{overpic}
	\end{subfigure}
	\caption{Aspects of mapping a subarctic boreal forest. (\texttt{II}) shows snow fall caused by the snow in the tree branches.  (\texttt{I}) was taken in \texttt{skidoo}, (\texttt{II}) in \texttt{lake}, and (\texttt{III})-(\texttt{IV}) in \texttt{forest}.
	}
	\label{fig:picture-of-issues}
\end{figure}
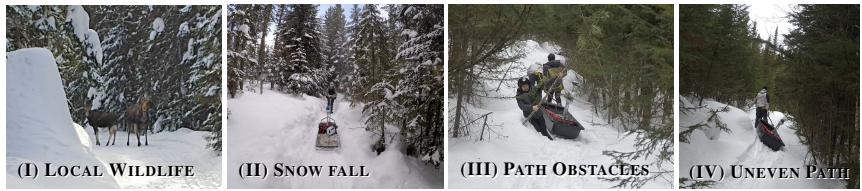
\textbf{\texttt{Lake}} -- In this dataset, the baseline map looks similar to the penalty map when observed from top.
However, a side view clearly shows that the buildings at the start and the end of the trajectory do not match.
This effect is avoided by applying the penalty.
For each configuration, we highlighted a pine tree in the top part of the map.
While the tree is crisper in the baseline, the penalty map's tree is clearly sharper than the one from the prior map.
\textbf{\texttt{Forest}} -- The particularity of this map is the rough trail at the bottom of the map (see subfig. \texttt{III} and \texttt{IV} of \autoref{fig:picture-of-issues}).
It suffers from large circular artifacts caused by the platform being immobile and by major changes of orientation along that trail.
Again, the penalty is less clear than the baseline, but quite an improvement compared to the prior map.
The circular-shaped building at the top part of the map is quite blurry in both the prior and penalty map.
This lack of crispiness is caused by the start and end of the trajectory not correctly matching.
\textbf{\texttt{Skidoo}} -- The baseline map is the most bent of all three datasets.
This bending has two probable causes: 1) the map was created with lower density $\epsilon = $ \SI{10}{\centi\meter} contrary to the other maps 2) the trajectory is twice as long as the \texttt{lake} dataset.
Otherwise, the behavior is similar to that of the other datasets with the similar circular-shaped artifacts in the prior map.

None of the baseline maps manages to close the loop, they all drift and bend over time.
Because of the magnetometer-based heading, the prior maps show large circular artifacts at several locations.
These artifacts are especially noticeable in zones where the platform stops moving---the heading cannot be corrected by \ac{GNSS} in this case.
All penalty and prior maps closed the loop in a similar fashion.
Moreover, the penalties manage to achieve a trade-off between the global and local consistency.

The field trials at the For\^{e}t Montmorency presented a number of challenges.
As \texttt{(I)} of \autoref{fig:picture-of-issues} shows, local wildlife might hinder your experiments.
We had a pair of moose blocking one of our trajectories and the experiment had to be rescheduled for another day.
Also, wild birds used our static \ac{GNSS} antenna as a perch.
Another challenge is snow fall, as even a light snow fall will be  visible in the map.
Furthermore, even when our experiments were done on a clear day, they were still affected by snow falling from the trees (see subfig. \texttt{II} of \autoref{fig:picture-of-issues}).
In the \texttt{forest} trajectory, we had to pass through an untapped trail with trees blocking our way \texttt{(III)}.
Because of the roughness of that trail, the platform almost tumbled over multiple time \texttt{(IV)}.
The snowmobile trails, on the other hand, were easy to pass through.

Cold can cause hindering issues with lidars that are not properly rated for low temperatures.
We tested a lidar rated for a minimum of \SI{-10}{\celsius}, with exterior temperature during our field experiments varying between \SI{-17}{\celsius} and \SI{-7}{\celsius}.
The lidar started to malfunction when the temperature dropped below \SI{0}{\celsius}, producing spurious measurements to a level where the environment could not be seen anymore.
Providing an exterior source of heat could mitigate the problem temporary.%
A second lidar rated for \SI{-20}{\celsius} was used for our final experiments.
Through the development process, we observed that sun glare was more apparent at low temperatures.
More tests are required to fully understand the impact of cold on lidar, but one should be careful regarding the temperature ranting of sensors deployed as it can cause serious safety issues.

\vspace{-12pt}
\subsection{Steps towards Real-Time Large-Scale Lidar Mapping}
\label{sec:res_realtime}
\vspace{-7pt}

\begin{SCfigure}
	\vspace{-10pt}
	\includegraphics[width=0.68\textwidth]{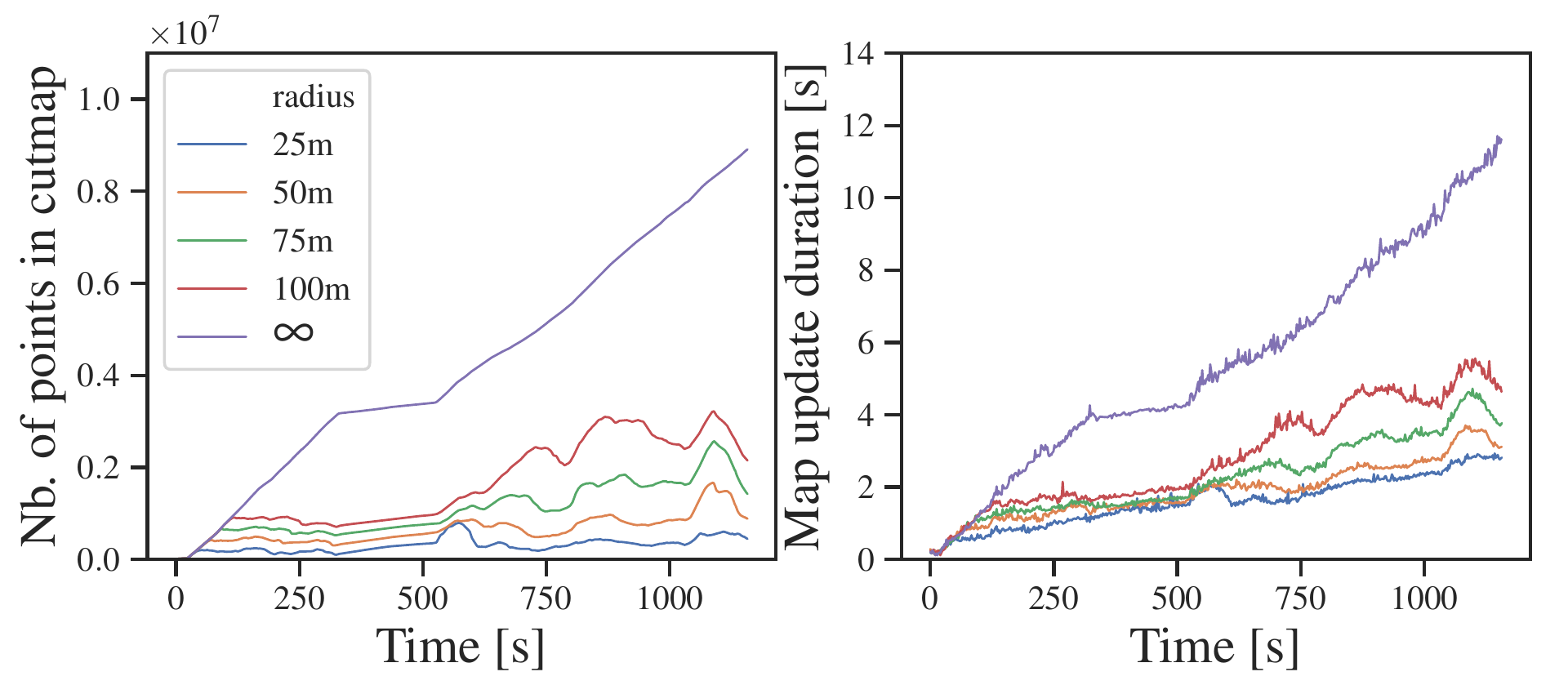}
	\caption{
		Effect of the cut-map radius $r_{\text{max}}$ on the complexity over time for the \texttt{lake} trajectory.
		\emph{Left}: number of points used as a reference point cloud for the \ac{ICP}.
		\emph{Right}: computation time used by the mapper for each new lidar scan.
	}
	\label{fig:cutmap_perf}
\end{SCfigure}

The large-scale point cloud maps bear several problems that complicate real-time deployment on autonomous vehicles.
The processing time required to register the scans and update the map may limit the agility of the vehicle.
Moreover, memory management needs to be taken into consideration.
For example, the $\sim$\num{23700000} data points of the \texttt{forest} final prior map consumed \SI{1.8}{\giga\byte} when stored in RAM.
To improve the mapping speed, we have implemented the $r_{\text{max}}$ cut-map radius as defined in \autoref{subsec:mapping}.
As shown in \autoref{fig:cutmap_perf}, the mapper execution time is reduced.
One can observe a flattening in the number of points in the cut-map around \SI{250}{\second} to \SI{500}{\second}.
This situation occurs when the mobile platform is immobile.

Finally, in order to achieve globally consistent maps, we used the \ac{RTK} \ac{GNSS} solution consisting of two receivers, one static and the other attached to the mobile platform.
The precise positioning information is obtained by combining the information from both receivers.
In our case, it was done during post-processing of the dataset.
For real-time deployment, it is necessary to reliably transmit the static receiver information to the mobile vehicle, which may be difficult in dense vegetation.

\vspace{-25pt}
\section{Conclusion}
\vspace{-14pt}
\label{sec:Conclusion}
In this paper, we explored the process of creating large globally consistent maps of subarctic boreal forests by adding external constraints to the \ac{ICP} algorithm.
The maps remained crisp even through \SI{4.1}{\kilo\meter} of narrow walkable and snowmobile trails.
We also discussed problems encountered with the environment and the lidar sensor during the field trials.
Moreover, we introduced a computation optimization for very large maps, allowing real-time deployments.
Encouraged by the results, this opens the door to further comparison with \ac{NDT} and \ac{GICP} using better experimental validation and external tracking systems.
Furthermore, studying the impact of penalties within \ac{ICP} against graph minimization would lead to a better understanding of their pros and cons.

\vspace{-8pt}
\begin{acknowledgement}
This works was financed by the Fonds de Recherche du Qu\'{e}bec -- Nature et technologies (FRQNT) and the Natural Sciences and Engineering Research Council of Canada (NSERC) through the grant CRDPJ 527642-18 SNOW.
\end{acknowledgement}
\vspace{-35pt}
\bibliography{references}

\begin{thebibliography}{27}
\providecommand{\natexlab}[1]{#1}
\providecommand{\url}[1]{\texttt{#1}}
\expandafter\ifx\csname urlstyle\endcsname\relax
  \providecommand{\doi}[1]{doi: #1}\else
  \providecommand{\doi}{doi: \begingroup \urlstyle{rm}\Url}\fi

\bibitem[Zlot and Bosse(2014)]{Zlot2014b}
R.~Zlot and M.~Bosse.
\newblock Efficient large-scale three-dimensional mobile mapping for
  underground mines.
\newblock \emph{JFR}, 31\penalty0 (5):\penalty0 758--779, 2014.

\bibitem[Pierzcha\l{}a et~al.(2018)Pierzcha\l{}a, Gigu{\`{e}}re, and
  Astrup]{Pierzchala2018}
M.~Pierzcha\l{}a, P.~Gigu{\`{e}}re, and R.~Astrup.
\newblock Mapping forests using an unmanned ground vehicle with {3D} {LiDAR}
  and graph-{SLAM}.
\newblock \emph{Computers and Electronics in Agriculture}, 145:\penalty0
  217--225, 2018.

\bibitem[Williams et~al.(2012)Williams, Parker, and Howard]{Williams2012}
S.~Williams, L.~T. Parker, and A.~M. Howard.
\newblock {Terrain Reconstruction of Glacial Surfaces : Robotic Surveying
  Techniques}.
\newblock \emph{RA Magazine}, 19\penalty0 (4):\penalty0 59--71, 2012.

\bibitem[Besl and McKay(1992)]{Besl1992}
P.~J. Besl and N.~D. McKay.
\newblock {A method for registration of 3-D shapes}.
\newblock \emph{TPAMI}, 14\penalty0 (2):\penalty0 239--256, 1992.

\bibitem[Pomerleau et~al.(2013)Pomerleau, Colas, Siegwart, and
  Magnenat]{Pomerleau2013}
F.~Pomerleau, F.~Colas, R.~Siegwart, and S.~Magnenat.
\newblock {Comparing ICP variants on real-world data sets: Open-source library
  and experimental protocol}.
\newblock \emph{Autonomous Robots}, 34\penalty0 (3):\penalty0 133--148, 2013.

\bibitem[Thrun and Montemerlo(2006)]{Thrun2006}
S.~Thrun and M.~Montemerlo.
\newblock {The GraphSLAM Algorithm with Applications to Large-Scale Mapping of
  Urban Structures}.
\newblock \emph{IJRR}, 25\penalty0 (5-6):\penalty0 403--429, 2006.

\bibitem[Cadena et~al.(2016)Cadena, Carlone, Carrillo, Latif, Scaramuzza,
  et~al.]{Cadena2016}
C.~Cadena, L.~Carlone, H.~Carrillo, Y.~Latif, D.~Scaramuzza, et~al.
\newblock {Past, Present, and Future of Simultaneous Localization and Mapping:
  Toward the Robust-Perception Age}.
\newblock \emph{T-RO}, 32\penalty0 (6):\penalty0 1309--1332, 2016.

\bibitem[Landry et~al.(2019)Landry, Pomerleau, and Gigu{\`{e}}re]{Landry2019}
D.~Landry, F.~Pomerleau, and P.~Gigu{\`{e}}re.
\newblock {CELLO-3D: Estimating the Covariance of ICP in the Real World}.
\newblock In \emph{ICRA}, 2019.

\bibitem[Lever et~al.(2013)Lever, Delaney, Ray, Trautmann, Barna, and
  Burzynski]{Lever2013}
J.~H. Lever, A.~J. Delaney, L.~E Ray, E.~Trautmann, L.~A. Barna, and A.~M.
  Burzynski.
\newblock {Autonomous GPR Surveys using the Polar Rover Yeti}.
\newblock \emph{JFR}, 30\penalty0 (2):\penalty0 194--215, 2013.

\bibitem[Jagbrant et~al.(2015)Jagbrant, Underwood, Nieto, and
  Sukkarieh]{Jagbrant2015}
G.~Jagbrant, J.~P. Underwood, J.~Nieto, and S.~Sukkarieh.
\newblock {LiDAR Based Tree and Platform Localisation in Almond Orchards}.
\newblock In \emph{FSR}, pages 469--483, 2015.

\bibitem[Williams and {M. Howard}(2009)]{Williams2009}
S.~Williams and A.~{M. Howard}.
\newblock {Developing Monocular Visual Pose Estimation for Arctic
  Environments}.
\newblock \emph{JFR}, 71\penalty0 (5):\penalty0 486--494, 2009.

\bibitem[Paton et~al.(2016)Paton, Pomerleau, and Barfoot]{Paton2016a}
M.~Paton, F.~Pomerleau, and T.D. Barfoot.
\newblock In the dead of winter: Challenging vision-based path following in
  extreme conditions.
\newblock \emph{STAR}, 113:\penalty0 563--576, 2016.

\bibitem[McDaniel et~al.(2012)McDaniel, Takayuki, Brooks, Salesses, and
  Lagnemma]{McDaniel2012}
M.~W. McDaniel, N.~Takayuki, C.~A. Brooks, P.~Salesses, and K.~Lagnemma.
\newblock {Terrain Classification and Identification of Tree Stems Using
  Ground-based LiDAR}.
\newblock \emph{JFR}, 29:\penalty0 891--910, 2012.

\bibitem[Wallace et~al.(2016)Wallace, Lucieer, Malenovský, Turner,
  et~al.]{Wallace2016}
L.~Wallace, A.~Lucieer, Z.~Malenovský, D.~Turner, et~al.
\newblock {Assessment of Forest Structure Using Two UAV Techniques: A
  Comparison of Airborne Laser Scanning and Structure from Motion (SfM) Point
  Clouds}.
\newblock \emph{Forests}, 7\penalty0 (3), 2016.

\bibitem[Tian et~al.(2013)Tian, P., d'Angelo, and Ehlers]{Tian2013}
J.~Tian, Reinartz P., P.~d'Angelo, and M.~Ehlers.
\newblock Region-based automatic building and forest change detection on
  cartosat-1 stereo imagery.
\newblock \emph{ISPRS}, 79:\penalty0 226 -- 239, 2013.

\bibitem[Geiger et~al.(2013)Geiger, Lenz, Stiller, and Urtasun]{Geiger2013IJRR}
A.~Geiger, P.~Lenz, C.~Stiller, and R.~Urtasun.
\newblock Vision meets robotics: The {KITTI} dataset.
\newblock \emph{Int. J. Robotics Res.}, 2013.

\bibitem[Pomerleau et~al.(2015)Pomerleau, Colas, and Siegwart]{Pomerleau2015b}
F.~Pomerleau, F.~Colas, and R.~Siegwart.
\newblock {A Review of Point Cloud Registration Algorithms for Mobile
  Robotics}.
\newblock \emph{Found. and Trends in Rob.}, 4\penalty0 (1):\penalty0 1--104,
  2015.

\bibitem[Ohta and Kanatani(1998)]{Ohta1998}
N.~Ohta and K.~Kanatani.
\newblock {Optimal Estimation of Three- Dimensional Rotation and Reliability
  Evaluation}.
\newblock \emph{ECCV}, pages 175--187, 1998.

\bibitem[Est{\'{e}}par et~al.(2004)Est{\'{e}}par, Brun, and
  Westin]{Estepar2004}
R.~S.~J. Est{\'{e}}par, A.~Brun, and C.-F. Westin.
\newblock {Robust Generalized Total Least Squares Iterative Closest Point
  Registration}.
\newblock In \emph{MICCAI}, 2004.

\bibitem[Maier-hein et~al.(2012)Maier-hein, Franz, Santos, Schmidt, Fangerau,
  et~al.]{Maier-hein2012}
L.~Maier-hein, A.~M. Franz, T.~R. Santos, M.~Schmidt, M.~Fangerau, et~al.
\newblock {Convergent \ac{ICP} Algorithm to Accomodate Anisotropic and
  Inhomogenous Localization Error}.
\newblock \emph{TPAMI}, 34\penalty0 (8):\penalty0 1520--1532, 2012.

\bibitem[Billings et~al.(2015)Billings, Boctor, and Taylor]{Billings2015}
S.~D. Billings, E.~M. Boctor, and R.~H. Taylor.
\newblock {Iterative Most-Likely Point Registration ( {IMLP} ): A Robust
  Algorithm for Computing Optimal Shape Alignment}.
\newblock \emph{PLoS ONE}, pages 1--45, 2015.

\bibitem[Segal et~al.(2009)Segal, Haehnel, and Thrun]{Segal2009}
A.~V. Segal, D.~Haehnel, and S.~Thrun.
\newblock {Generalized-{ICP}}.
\newblock In \emph{RSS}, 2009.

\bibitem[Chen and Medioni(1992)]{Chen1992}
Y.~Chen and G.~Medioni.
\newblock {Object modeling by registration of multiple range images}.
\newblock In \emph{ICRA}, pages 2724--2729, 1992.

\bibitem[Babin et~al.(2019)Babin, Gigu{\`{e}}re, and Pomerleau]{Babin2019}
P.~Babin, P.~Gigu{\`{e}}re, and F.~Pomerleau.
\newblock {Analysis of Robust Functions for Registration Algorithms}.
\newblock In \emph{ICRA}, 2019.

\bibitem[Balachandran and Fitzpatrick(2009)]{Balachandran2009}
R.~Balachandran and J.~M. Fitzpatrick.
\newblock {Iterative Solution for Rigid-Body Point-Based Registration with
  Anisotropic Weighting}.
\newblock \emph{SPIE}, 7261:\penalty0 1--10, 2009.

\bibitem[Pomerleau et~al.(2014)Pomerleau, Kr{\"{u}}si, Colas, Furgale, and
  Siegwart]{Pomerleau2014}
F.~Pomerleau, P.~Kr{\"{u}}si, F.~Colas, P.~Furgale, and R.~Siegwart.
\newblock {Long-term 3D map maintenance in dynamic environments}.
\newblock In \emph{ICRA}, pages 3712--3719, 2014.

\bibitem[Laconte et~al.(2019)Laconte, Desch\^{e}nes, Labussi\`{e}re, and
  Pomerleau]{Laconte2019}
J.~Laconte, S.-P. Desch\^{e}nes, M.~Labussi\`{e}re, and F.~Pomerleau.
\newblock Lidar measurement bias estimation via return waveform modelling in a
  context of 3d mapping.
\newblock In \emph{ICRA}, 2019.

\end{thebibliography}

\end{document}